\documentclass{article}

\PassOptionsToPackage{numbers, compress}{natbib}
     \usepackage[final]{neurips_2021}

\usepackage{enumitem}
\usepackage{xcolor}
\usepackage{multirow}
\usepackage{array}
\usepackage{siunitx}

\usepackage[utf8]{inputenc} % allow utf-8 input
\usepackage[T1]{fontenc}    % use 8-bit T1 fonts
\usepackage{hyperref}       % hyperlinks
\usepackage{url}            % simple URL typesetting
\usepackage{booktabs}       % professional-quality tables
\usepackage{amsfonts}       % blackboard math symbols
\usepackage{nicefrac}       % compact symbols for 1/2, etc.
\usepackage{microtype}      % microtypography
\usepackage{xcolor}         % colors

\usepackage{graphicx}
\title{Stakeholder Participation in AI:\\Beyond ``Add Diverse Stakeholders and Stir''}

\author{%
  Fernando Delgado \\
  Cornell University \\
   \texttt{fad33@cornell.edu} \\
  \And
  Stephen Yang \\
  Cornell University \\
  \texttt{sy364@cornell.edu} \\
    \AND
  Michael Madaio \\
  Microsoft Research \\
  \texttt{michael.madaio@microsoft.com} \\
    \And
  Qian Yang \\
  Cornell University \\
  \texttt{qianyang@cornell.edu} \\

}

\begin{document}

\maketitle

% \begin{abstract}
%   \input{abstract}
% \end{abstract}

\section{Introduction}

% HCI and AI are both interested in end-user and stakeholder participation as a way to make AI better respresent values and ensure ethics.
Both AI and HCI communities increasingly call for more stakeholder participation in AI system design, development, and maintenance \cite{icml2020wkshp,informs2021wkshp,loi2018pd,cscw2018wkshp,pair2020boundary}.
Participation can allow AI systems to better reflect their end users' and stakeholders’ values, preferences, and needs. It can also help scholars and practitioners to better anticipate and account for AI's negative downstream impacts, such as fairness and equity breakdowns \cite{choudhury2020introduction,algo_2021, hoffmann2020terms,deloitte2018participatory}.
In parallel and responding to these calls, participatory AI projects and research publications have started to emerge across HCI and AI communities \cite{robertson2020if,martin2020participatory,krafft2021facct}.

% (1) Seemingly all participation counts. (2) Unclear how to think about levels of participation. 
% (3)No principled discussion.
Despite the growing consensus that end-users and stakeholders \textit{should} participate, enormous variation and implicit disagreements exist among current approaches to participatory AI.
For example, researchers frequently described their approach as ``participatory design workshops'', yet there is a noticeable lack of discussion around the tactics deployed in these workshops~\cite{lee2019webuildai,smith2020keeping}. For example, how did the workshops/interviews account for stakeholders' diverse (and sometimes conflicting) values? Whether or how were they more effective than focus groups and interviews in traditional user-centered design or agile development processes? %; how AI researchers and designers translated stakeholders' high-level fairness inputs into numerous data processing or modeling decisions that shape the AI concretely.
For AI practitioners who are interested in taking a participatory approach to AI design/development, it remains challenging to assess the extent to which a participatory tactic or process can actually achieve inclusiveness or fairness goals.
For researchers who are interested in advancing this participatory turn in AI, it remains challenging to have a principled discussion about the pros and cons of existing approaches and how we might best choose to do so in going forward \cite{does-AI-make-PD-obsolete}.

% method paragraph: We have expertise in PD and HAI design. We have studied the field of PD/AI. With these expertise and experiences, we bring rigor to PD/AI discussion.
% As this workshop's mission states: Values and ethics are necessarily entangled with localized, situated, and culturally-informed human perspectives.
This paper aims to add theoretical grounding, structure, and clarity to the emergent research discourses around stakeholder participation in AI.
Taking a lesson from early HCI work in integrating women’s perspectives into a male-dominated curriculum, we cannot just add diverse end-users and stakeholders ``\textit{and stir. It takes work, new ways of thinking, and new kinds and methods of openness, to bring substantively new voices into a conversation.}''~\cite{PD-Muller-2002}
This paper takes this proposition as a starting place. We first offer a brief overview of the many approaches to increasing participation elaborated in HCI design, political theory, and social science research. We derive \textit{five dimensions of participation} that AI practitioners or researchers can readily use in assessing the extent to which any participatory tactic or process meaningfully empowers stakeholders in AI design. 
Finally, we highlight three challenges that practitioners face when taking participatory approaches to AI design.
These findings come from our analysis of 56 research publications on participatory AI, as well as 12 IRB-approved semi-structured interviews with researchers and practitioners who had written one or more of these publications.
With these theoretical and empirical analyses, we hope to push forward a principled discussion around stakeholder participation in AI design as a way to account for diverse human values and ethics.% -- as the proposal of this workshop states -- ``\textit{are necessarily entangled with localized, situated, and culturally-informed human perspectives.}".

\section{Navigating the Many Approaches to Stakeholder Participation}

% Approaches to participation in HCI
% Efforts to increase stakeholder participation in AI most often referenced participatory design (PD) \cite{simonsen2012routledge,PD-Muller-2002} as their rhetorical or theoretical inspiration.
% However, this body of work typically did not describe how their participation tactics adhered to or deviated from PD traditions. Some used ill-defined terms that only gestured at PD (e.g., keeping stakeholders in the loop). Others used the terms participatory design, co-design and value-sensitive design interchangeably. 
This paper aims to provide more clarity and structure to the research discourses around participatory AI. As a first step, we review the range of existing participatory approaches across HCI design, political theory, and social science research.
We first highlight their differences, and then identify five questions that these diverse approaches address (namely, five ``dimensions of participation'').

\subsection{Prior Approaches to Participation and Their Differences}

\begin{itemize}[leftmargin=*]
    \item \textbf{User-centered design:} inquiring stakeholders during need finding and design evaluation.

    Over the past decades, the double-diamond user-centered design process has become the most common design process in industry practices \cite{doublediamond}. It actively engages with end users when technology designers work to identify user needs and assess their design ideas \cite{mao2005state,olson2014ways}. 

    \item \textbf{Service design:} understanding and shaping value propositions and interactions among stakeholders.
    
    Service design shares many features with user-centered design, but expands the notion of the ``users'' to other stakeholders impacted by the technology service \cite{forlizzi2018moving, forlizzi2013promoting,forlizzi2018moving}. Using methods such as \textit{service blueprint} and \textit{value stream mapping}, service designers explicitly consider how their design decisions impact of the constellation of stakeholders through explicit interactions or implicit value propositions over time. 
    Some scholars consider service design as ``participatory'' \cite{saad2020service,holmlid2012participative}. 
    
    \item \textbf{Participatory design (PD):} mixing diverse voices and challenging power structures.
    
    In user-centered design and service design, technology designers collect, synthesize, and translate stakeholders' diverse inputs into concrete design decisions. In contrast, PD aims to enable stakeholders to provide direct inputs in technology design \cite{gregory2003scandinavian,simonsen2012routledge}.
    For stakeholders--who typically are not experts in the technology themselves--to provide valuable inputs, PD scholars highlight that participation needs to ``\textit{take place neither in the users’ domain nor in the technology developers’ domain}'' \cite{muller1993participatory}. PD practitioners need to create a hybrid space that allows diverse stakeholders to meaningfully contribute ``\textit{a mix of motivations, histories, and goals}'' without a clear set of ``\textit{authority relations, incentives, and obligations}''
    \cite{le_dantec_strangers_2015,harrington2019deconstructing}. Noteworthily, PD highlights the work of uncovering and challenging existing ``authority relations'', including both power structures between technology designers and impacted stakeholders, as well as power dynamics among and within each group of impacted stakeholders \cite{muller1993participatory}.
    % Qian: to-do: add examples.

    \item \textbf{Co-design:} creative cooperation between stakeholders and technology designers across the whole span of a design process \cite{def-codesign-sanders,co-design-as-joint-inquiry-imagination,def-codesign-Kleinsmann}. While closely related to PD, co-design typically lacks its explicitly political component \cite{co-design-as-joint-inquiry-imagination}. %Further, co-design often focuses more explicitly on designing specific products or artifacts \cite{def-codesign-Kleinsmann}, while PD may focus more on the larger sociotechnical system in which particular artifacts will be embedded \cite{simonsen2012routledge}.
    
    % Co-design has many synergies with action research.
    \item \textbf{Action research:} improving researchers' capacities and processes \cite{elliot1991action}. Particularly relevant to this work is \textit{participatory} action research approaches, which engage stakeholders as co-inquirers and co-construct research plans and interventions \cite{unertl2016integrating,wallerstein2006using,hayes2014knowing, rasmussen2004action}.
    This normative aim to improving research processes differentiates participatory action research from PD, despite their many similarities \cite{gleerup2019action}. Some recent HCI research has begun to adopt action research approaches \cite{hayes2014knowing, harrington2019deconstructing}.

    % (e.g., teachers' reflexive AR \cite{elliot1991action}), although more participatory approaches to AR (i.e., PAR) \cite{unertl2016integrating,wallerstein2006using} engage other stakeholders as ``co-inquirers'' who ``co-construct'' research plans and interventions with researchers~\cite{hayes2014knowing, rasmussen2004action}.
    % While similar to PD, participatory action research focuses more on collaboratively designing the research process or improving practitioners' or communities' capacities. Some HCI research has begun to adopt both AR \cite{hayes2014knowing} and PAR approaches~\cite{harrington2019deconstructing}.
    % \textbf{Action Research:}, originating from the social sciences, focuses on facilitating democratic participation at the community-level. AR was developed to bring about actions ``\textit{'with' people experiencing real problems in their everyday lives not 'for,' 'about,' or 'focused on' them}'' \cite{hayes2014knowing}. %AR is informed by a ``\textit{critical normative aim to contribute to democratic development of society through empowering people to take part in the discussions and decision-making processes that surround and determine their lives}'' \cite{gleerup2019action}.
    % The AR approach to participatory democracy place individuals as "co-inquirers" that work with researchers in a collaborative relationship to reflect on their own experiences and "co-construct" actions and interventions \cite{hayes2014knowing, rasmussen2004action}. \michael{maybe a line about how this approach is beginning to be adopted in HCI e.g., \cite{harrington2019deconstructing}?}
    
    \item \textbf{Value-sensitive design:} accounting for indirect stakeholders and moral values.
    Value-sensitive design can seem similar to participatory design, yet with a particular emphasis on the ethical values of direct and indirect stakeholders \cite{friedman1996value,friedman2002value}. While valuable and appealing, scholars have argued that value-sensitive design broadens the scope of PD to an unmanageably large scale \cite{borning2004designing}.

    \item \textbf{Social choice theory and mechanism design:} quantitative aggregation of stakeholder preferences.
    
    Social choice theory focuses on identifying individuals' preferences and developing an aggregated, mathematical preference-ranking model (for example, polling and ranking individuals' public policy preferences) \cite{sen1977social,arrow2012social,sen1986social}.
    In recent years, the field of mechanism design has translated this approach into a framework for including stakeholders into algorithmic decision-making \cite{abebe2018mechanism,finocchiaro2021bridging,hitzig2020normative,viljoen2021design}.
    
    \item \textbf{Participatory democracy and civic participation:} involving citizens and stakeholders in a broad range of civic decision making.
    % while sharing certain concerns regarding social welfare with SCT, have a longer lineage spanning several decades \cite[e.g.,][]{arnstein1969ladder,fung2006varieties,polletta2012freedom, lippmann1993phantom, dewey19541927, macpherson_democratic_1973}, and focus on issues concerning political action and civic decision-making, making them popular with social movements \cite{polletta2012freedom}.
    Participatory democracy conceptualizes civic participation as a spectrum, from least intense (e.g., stakeholders spectating or expressing preferences) to most intense (e.g., deliberating, negotiating, or deploying expertise through dialogues) \cite{arnstein1969ladder,fung2006varieties,polletta2012freedom, lippmann1993phantom, dewey19541927, macpherson_democratic_1973}.\looseness=-1 
    
    \item \textbf{Deliberation theory:}~qualitatively weighing and discussing competing perspectives and policies.
    Deliberation theory emerged in response to mechanistic approaches to aggregating stakeholder preferences (e.g., social choice theory) \cite{oberg2016deliberation,fishkin2005experimenting,owen2015survey}. It emphasizes the importance of bringing together small groups of people to discuss and \textit{qualitatively} weigh competing arguments for policies \cite{fishkin2005experimenting}.\looseness=-1
    
\end{itemize}

% Separately, originating from the social sciences, the tradition of Action Research (AR) is focused on facilitating democratic participation at the community-level. AR was developed to bring about actions ``\textit{'with' people experiencing real problems in their everyday lives not 'for,' 'about,' or 'focused on' them}'' \cite{hayes2014knowing}. AR is informed by a ``\textit{critical normative aim to contribute to democratic development of society through empowering people to take part in the discussions and decision-making processes that surround and determine their lives}'' \cite{gleerup2019action}. The AR approach to participatory democracy place individuals as "co-inquirers" that work with researchers in a collaborative relationship to reflect on their own experiences and "co-construct" actions and interventions \cite{hayes2014knowing, rasmussen2004action}. \michael{maybe a line about how this approach is beginning to be adopted in HCI e.g., \cite{harrington2019deconstructing}?}

\subsection{Dimensions of Participation: An Analytical Framework}

\begin{figure} [t]
  \includegraphics[width=0.95\linewidth]{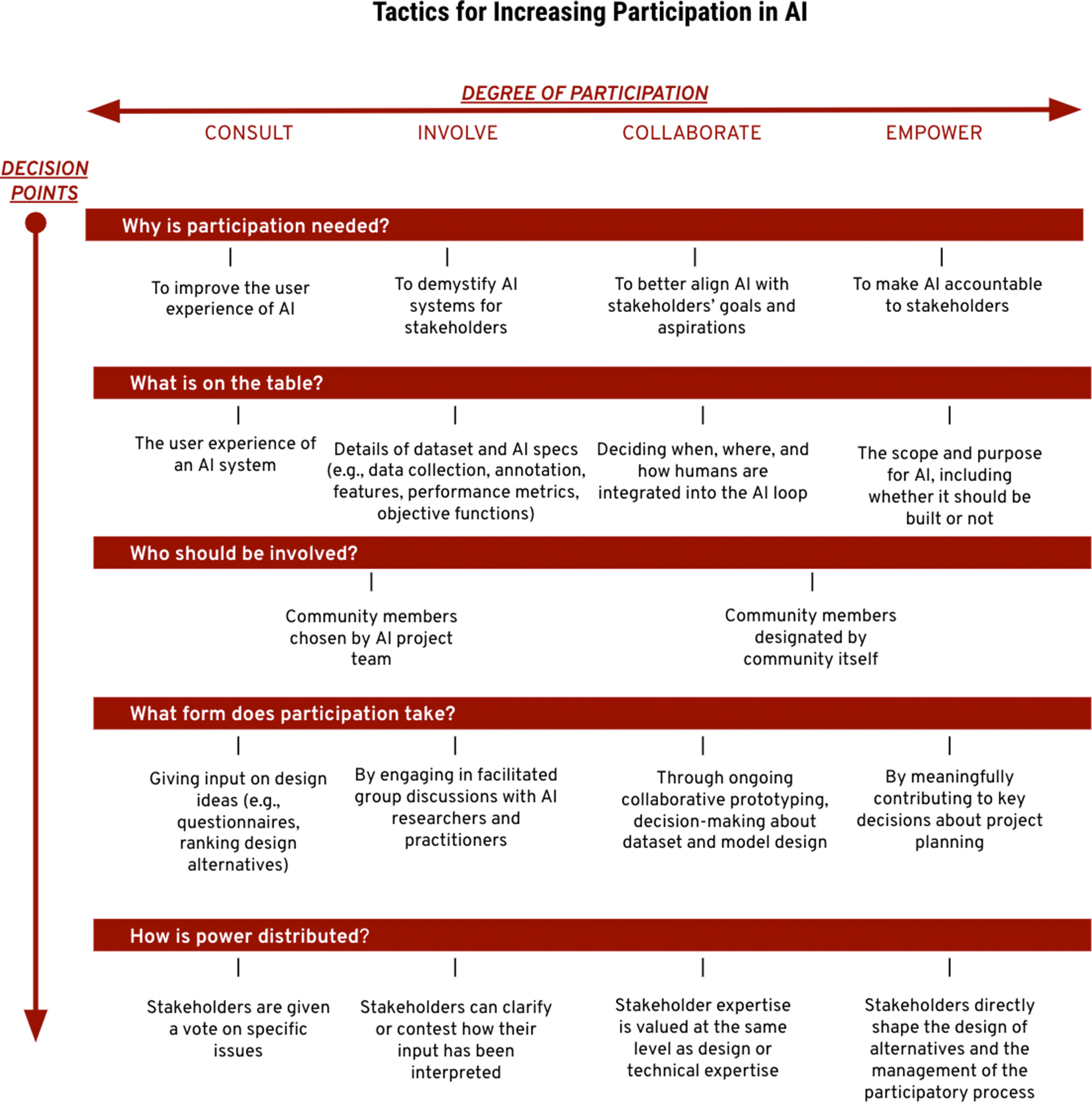}
  \caption{Tactics for increasing stakeholder participation in designing AI, from consulting to empowering. These are sequential decisions that AI practitioners and researchers need to make when including stakeholders in their AI design and development decision-making.}%Based on our literature synthesis and researcher/practitioner interviews, we identify a set of sequential decisions that AI practitioners and researchers need to make when including stakeholders in their AI design and development decision-making.}
  \label{fig:TIP}
\end{figure}

From the existing approaches to participation, we identify five questions that they collectively address: Why is participation needed? What is on the table? Which stakeholders should be involved? What form does their participation take? Finally (although this is cross-cutting across the other questions), how is power distributed among the participating stakeholders and between stakeholders and technology designers/engineers?
These five questions can serve as a valuable analytical tool for AI practitioners and researchers. They can help illuminate the often implicit differences across the aforementioned existing approaches to and theories of participation. These questions also aid practitioners in assessing the extent to which a chosen participatory tactic or process can indeed meaningfully empower diverse stakeholders in their AI design (Figure \ref{fig:TIP}). 

Take ``what form does participation take'' as an example.
If taking a \textit{Social Choice Theory} approach, AI practitioners can poll stakeholders and mathematically aggregate and rank their preferences afterwards~\cite{robertson2020if,hitzig2020normative}.
Advocates of \textit{Participatory Design} traditions might argue that this polling-and-preference-aggregation approach fails to account for the power imbalance between AI researchers and stakeholders. The power to translate stakeholders' high-level values and preferences into concrete AI design decisions still lies in the hands of the researchers \cite{muller2002participatory}.
\textit{Participatory democracy} research offers additional alternative forms of participation, for example, AI researchers can bring together stakeholders to deliberate merits and negotiate trade-offs of particular design decisions. They can encourage stakeholders to provide evidence to support their policy preferences and to change their preferences in light of new information as emphasized by advocates of \textit{Deliberation Theory} \cite{oberg2016deliberation,fishkin2005experimenting, fishkin2017prospects}.
By synthesizing how different theories of participation addressed this question of ``what form does participation take'', researchers and practitioners can better consider a wide range of options available, with awareness of their known strengths and weaknesses.

% Proponents of deliberation theory argue that decision-making and policy development should involve a . For deliberation theorists, participants should brought together to discuss merits and tradeoffs of particular policies and  In some applications of SCT, people might participate by expressing preferences via a ranking of some alternative options, as in approaches used in matching children to schools based on preferences \cite{robertson2020if,hitzig2020normative}. 

% In PD traditions, participants are brought together with designers in ``hybrid spaces'' to bridge experiential gaps between designers' technical expertise and participants' lived expertise \cite{muller2002participatory}. These hybrid spaces often take the form of workshops where participants and designers will engage in mutual learning of each others' experiences through collective ``reflection-in-action'' \cite{simonsen2012routledge}. Participants might engage in a variety of activities, including storytelling, roleplay, co-creation of prototypes, and more, to move from the current state to a more ideal envisioned future state \cite{muller2002participatory,simonsen2012routledge}. Unlike traditional user-centered design, PD directly involves stakeholders in helping bridge the gap between what is and what could be, to enable participants to envision and realize potential futures \cite{muller1993participatory,simonsen2012routledge}.

\section{Mapping the Current Participatory Practices in Designing AI}

% Findings summary from the CHI section 5.0.
We have identified a range of approaches to stakeholder participation in design (of policies, research, and technology). We have also identified five dimensions with which AI practitioners can synthesize this wide range of approaches to participation. %thereby making thoughtful choices.
In this section, we report on challenges and tensions that AI practitioners face when making decisions along the dimensions we outline in Figure~\ref{fig:TIP}. %deciding what is on the table, which stakeholders to involve, what form their participation will take, and how to distribute power among the participating stakeholders and between stakeholders and technology designers/developers.
These tensions and challenges emerged from our analysis of participatory AI publications and interviews with AI researchers and practitioners, providing key topics for workshop discussion and future research in increasing stakeholder participation in AI.

\textbf{Organizational mandates substantively constrain the scope of what is on the table and who is involved.}
Interviewees across the industry, public sectors, and academia reported that top-down organizational constraints largely determined ``\textit{what is on the table}'' and ``\textit{which stakeholders are involved}'' in their participatory AI projects. Corporations', research teams', and government agencies' timelines, priorities, mandates, and resource levels are often in tension with their desire to maximally empower stakeholders. 
Multiple interviewees in the public sector, for example, described that legislative efforts mandated the use of AI risk assessment tools \cite[e.g.,][]{saxena2021framework, algo_2021}. Only \textit{how} - not  \textit{whether} - AI will be deployed is on the table for stakeholders in such cases. These observations raise crucial questions for the HCI and AI communities: How can our larger organizational structures ensure meaningful stakeholder participation, for instance, in deciding whether an AI should exist at all? How should AI practitioners manage the timespan and nature of diverse stakeholder participation, in the context of agile development cycles in industry and relatively short project and funding cycles in academia?\looseness=-1

\textbf{Participatory AI project owners exert executive authority in deciding tactics of participation.}
Despite their intentions to empower stakeholders and democratize the AI design process, those who ``own'' the participatory AI project had unparalleled authority in making decisions about participatory approaches in practice (including the questions we pose in Figure~\ref{fig:TIP}). They often, if not always, decided who are considered stakeholders, what role each stakeholder plays, how they interact, whether they need to reach a consensus at the end, etc. These observations urge us to discuss what a more democratic method might look like for deciding the approaches to stakeholder participation.

\textbf{AI researchers and practitioners feel caught between a desire to fulfill an idealized vision of empowered stakeholder participation and real-world constraint on time and resources.}
% both human and algorithmic proxies are leveraged by AI researchers and practitioners as a way to tactically broaden participation when it is deemed impossible to have direct stakeholder interaction on an ongoing basis.
Interviewees reported that they felt caught between an idealized, ambitious vision of stakeholder empowerment and the practical constraints of time and resources. % (P9, P10)
One interviewee described what they viewed as an idealized version of participation as ``\textit{an absurd argument...’’} because ``\textit{if I wanted full participation meaning like I want this person literally coming to the office with me and making every decision with me and doing all these things, all of a sudden, they don't have a life to live, right?}'' %[P10].
% This straw-man description of participation seems to serve at one level to justify a diluted version of stakeholder involvement, and another as a source of anxiety about never being able to do participation in design any justice. 
In light of the unrealistic vision of ``full participation'', participatory AI projects we surveyed made expansive use of proxies as stand-ins for broader classes of stakeholders who are brought in to provide a stakeholder voice to address discrete design challenges \cite[cf.][]{mulvin2021proxies}.
For example, a project on educational AI asked educators to ``\textit{think like they were [children's] personas}'', thereby helping to design the AI for children. In another project described in our interviews, a stakeholder participated in an AI design process by training a machine learning model of their value and preferences. This model will vote for a resource allocation outcome on their behalf in the future, for an unspecified amount of time, even if the stakeholder's preferences change.\looseness=-1 %Similarly, we heard from interviewees how they made use of advocate-like proxies, or ``\textit{local confederates}'' to tap into larger sets of stakeholders they as AI researchers and practitioners were not directly in contact with.

This tension between the ideal and pragmatic constraints of participation reveals crucial open research questions. On the one hand, what constitutes a minimal level of meaningful participation, for instance, when deciding whether or how human and algorithmic proxies might meaningfully represent the values and preferences of a group of stakeholders?
On the other, at what point do stakeholder participation and empowerment reach a point of diminishing returns? 
Addressing these questions is a vital next step in improving stakeholder participation in AI design and development practice.

% \begin{itemize}[leftmargin=*]
    % \item % For RQ2, we find in our analysis that 
    % These participatory practices, and the larger participatory design effort they are leveraged within, are shaped by: (a) top-down organizational mandates which specify who should and who should not be involved as a participant, and to what extent these participant stakeholders have any voice in redirecting the design process or technology vision; and (b) the discretion of the design experts leading or staffed on these projects, whose decisions are predicated in large part on what they perceive to be the abilities of their partners and participant stakeholders.
    
    % \item % As it relates to RQ3, 
    % What also emerges from our analysis is a varied landscape of consultative approaches that elicit preferences and values from stakeholders that are then analyzed by designers and researchers in a broader loop of traditional design activities. Participation in this vein begins once the scope of the project is set and the process for interaction has been defined. Rather than affording participant stakeholders the power to (co-)lead the design process, participation in the current state of AI design practice is generally speaking—though with some exceptions—not rooted in a traditional PD vein of transferring design decision-making authority to the participant stakeholders. In fact, in the majority of projects surveyed in this analysis, participation was operationalized as stakeholder involvement in one workshop session that lasted less than 3 hours.
% \end{itemize}

\clearpage

% \begin{ack}
% \end{ack}
% References follow the acknowledgments. 
%\bibliographystyle{unsrt}
\bibliographystyle{plain}
\bibliography{main}
\end{document}